\documentclass{article}

\usepackage{PRIMEarxiv}

\usepackage[utf8]{inputenc} % allow utf-8 input
\usepackage[T1]{fontenc}    % use 8-bit T1 fonts
\usepackage{hyperref}       % hyperlinks
\usepackage{url}            % simple URL typesetting
\usepackage{booktabs}       % professional-quality tables
\usepackage{amsfonts}       % blackboard math symbols
\usepackage{nicefrac}       % compact symbols for 1/2, etc.
\usepackage{microtype}      % microtypography
\usepackage{lipsum}
\usepackage{fancyhdr}       % header
\usepackage{graphicx}       % graphics
\graphicspath{{media/}}     % organize your images and other figures under media/ folder

\usepackage{cite}
\usepackage{amsmath,amssymb}
\usepackage{algorithmic}
\usepackage{graphicx, enumitem}
\usepackage{textcomp}
\usepackage{xcolor}

\title{Inherently Interpretable and Uncertainty-Aware Models for Online Learning in Cyber-Security Problems}

\author{
  Benjamin Kolicic, Alberto Caron, Chris Hicks, Vasilios Mavroudis \\
  The Alan Turing Institute \\
  London, UK\\
  \texttt{\{bkolicic, acaron, c.hicks, vmavroudis\}@turing.ac.uk} \\
}

\begin{document}
\maketitle

\begin{abstract}
In this paper, we address the critical need for interpretable and uncertainty-aware machine learning models in the context of online learning for high-risk industries, particularly cyber-security. While deep learning and other complex models have demonstrated impressive predictive capabilities, their opacity and lack of uncertainty quantification present significant questions about their trustworthiness. We propose a novel pipeline for online supervised learning problems in cyber-security, that harnesses the inherent interpretability and uncertainty awareness of Additive Gaussian Processes (AGPs) models. Our approach aims to balance predictive performance with transparency while improving the scalability of AGPs, which represents their main drawback, potentially enabling security analysts to better validate threat detection, troubleshoot and reduce false positives, and generally make trustworthy, informed decisions. This work contributes to the growing field of interpretable AI by proposing a class of models that can be significantly beneficial for high-stake decision problems such as the ones typical of the cyber-security domain. The source code is available \href{https://www.dropbox.com/scl/fo/ufcg39kki34vw6d6u5lwv/AIqL42zXlaKo7jdLGCEHPU0?rlkey=tl84jtfefrvc4z1mjjpfydk4c&st=98dij2dj&dl=0}{\textit{here}}.
\end{abstract}

% keywords can be removed

\section{Introduction}
Machine Learning (ML) models, particularly deep learning architectures, have achieved remarkable performance across various domains. However, their increasing complexity has led to concerns about their interpretability and the quantification of uncertainty in their predictions. Interpretability and explainability are desirable properties that refer to the ability to understand and explain the decision-making process of ML models~\cite{lipton2018mythos, molnar2020interpretable, arrieta2020explainable}. While uncertainty-quantification is necessary to assess the confidence of model prediction, enabling identification of regions of data where the model is less certain and may be prone to errors~\cite{abdar2021review}. Interpretability and uncertainty-awareness are thus pivotal features for building trust, ensuring fairness, and facilitating debugging and improvement of models, especially in high-stakes environments such as health-care~\cite{davenport2019potential}, cyber-security~\cite{berman2019survey, wang2020explainable, hicks2023canaries, bates2023reward} and finance~\cite{bracke2019machine}.

In cyber-security, the need for interpretable and uncertainty-aware ML models is particularly critical, due to the dynamic nature of the threats~\cite{buczak2015survey, hicks2024turing, sommer2010outside, thompson2024entity}. These are model features that enable security analysts to validate threat detection, reduce false positives, and prioritize alerts effectively~\cite{apruzzese2018effectiveness}. Interpretable models are also essential for regulatory compliance and legal contexts~\cite{goodman2017european}. Furthermore, such models provide insights to update defences against evolving threats~\cite{brundage2018malicious, mavroudis2023adaptive}, making them indispensable for robust and trustworthy cyber-security mechanisms. These benefits are realized when the model provides faithful explanations of its decision-making process.

The literature comprises works on ex-post explainability and inherent interpretability, a crucial distinction in machine learning, particularly in high-stakes domains like cyber-security~\cite{rudin2019stop,rudin2022interpretable}. Ex-post methods, such as SHAP~\cite{shapley:book1952,lundberg2017unified} and LIME~\cite{ribeiro2016should}, explain decisions of pre-trained models retrospectively, offering insights without restricting predictive power, but can suffer from inaccuracies and inconsistencies~\cite{rudin2019stop}. In contrast, inherently interpretable models, such as linear regression and decision trees, prioritize transparency from the outset, ensuring faithful and consistent explanations, though often at the expense of predictive performance. High-stakes environments like cyber-security benefit from coherent and faithful interpretations, making the choice of the interpretability approach a critical consideration.

Our work specifically addresses general online machine learning settings, which are less studied but prevalent in cyber-security problems. We showcase a class of models based on the theory of Generalized Additive Models~\cite{hastie2017generalized} called Additive Gaussian Processes (AGPs)~\cite{duvenaud2011additive, zhang2024gaussianprocessneuraladditive} that strike a good balance between predictive performance and inherent interpretability. However, despite these benefits and the trust they bring in high-stake scenarios, AGPs have never been applied to cyber-security problems and have severe scaling issues for large, feature-rich datasets in such online problems and the dynamic learning environment they bring. This scaling issue is perhaps the main restriction in uptake for such an informative model in the industry.

Through changes in the specific online learning pipeline of AGPs, discussed in later sections, we have managed to ease the scaling issue without incurring a significant loss in terms of model performance, as demonstrated by our empirical results in the experimental section.

\vspace{0.2cm}
 
\textbf{Contributions.} This paper demonstrates the advantages brought about by a class of Generalized Additive Models (GAMs)~\cite{hastie2017generalized} for cyber-security online learning problems. The GAM model variants we specifically take into account and assess are two: i) one based on neural nets, also known as Neural Additive Models (NAMs)~\cite{agarwal2021neural, radenovic2022neural, chang2022nodegam}; ii) the second one based on Additive Gaussian Processes (AGPs) \cite{williams2006gaussian, duvenaud2011additive, zhang2024gaussianprocessneuraladditive}. We showcase both models' properties in terms of their inherent interpretability of the regression/classification output, stemming from input-specific basis function modelling and their uncertainty-quantification properties, a task where AGPs thrive, while neural nets-based models struggle due to the notorious problem of over-confidence~\cite{guo2017calibration, lakshminarayanan2017simple, rahaman2021uncertainty}.

\section{Background}

We consider a general online supervised learning setting where the goal is to construct a learner based on the pair $(x_t, y_t) \in \mathcal{X} \times \mathcal{Y}$, that predicts the label $y_t = f(x_t)$, given a context realization $x_t \in \mathcal{X}$. We focus on classification problems where $y_t \in \{0, 1\}$, but the methods can be easily extended to regression ones as well. We assume, as it is common in many online cyber-security problems, a setting where at each step $t$: 
\begin{enumerate}[topsep=0pt,itemsep=0ex,partopsep=0ex,parsep=0ex, leftmargin=4ex]
    \item we receive an input sample $x_t \in \mathcal{X}$, or a batch of input samples $\{x_t\}^T_{t=1}$
    \item we predict the corresponding label via $\hat{y}_t = f(x_t)$
    \item we receive all the corresponding true labels $\{y_t\}^T_t$ (or just a portion of them)
    \item Finally, we update the learner $f(\cdot)$ based on a loss function $\ell (y_t, \hat{y}_t)$
\end{enumerate}
Note that in this online learning setup, samples are independent of each other, i.e., $(x_t, y_t) \perp \!\!\! \perp (x_{t+1}, y_{t+1})$. As a concrete example, this can represent a cyber-security problem where users attempt/request access to a service, one at a time. Users' accesses can be safely assumed to be independent of each other.

Before delving into the proposed interpretable models' structure, we briefly introduce Gaussian Processes models for regression/classification problems and their desirable uncertainty quantification properties \cite{williams2006gaussian}, and then discuss the intrinsic interpretability offered by the Generalized Additive Models (GAMs) formulation \cite{hastie2017generalized}. We then will illustrate how Additive Gaussian Processes cleverly combine these two properties together.

\begin{figure}[t]
    \centering
    \includegraphics[width=0.48\textwidth]{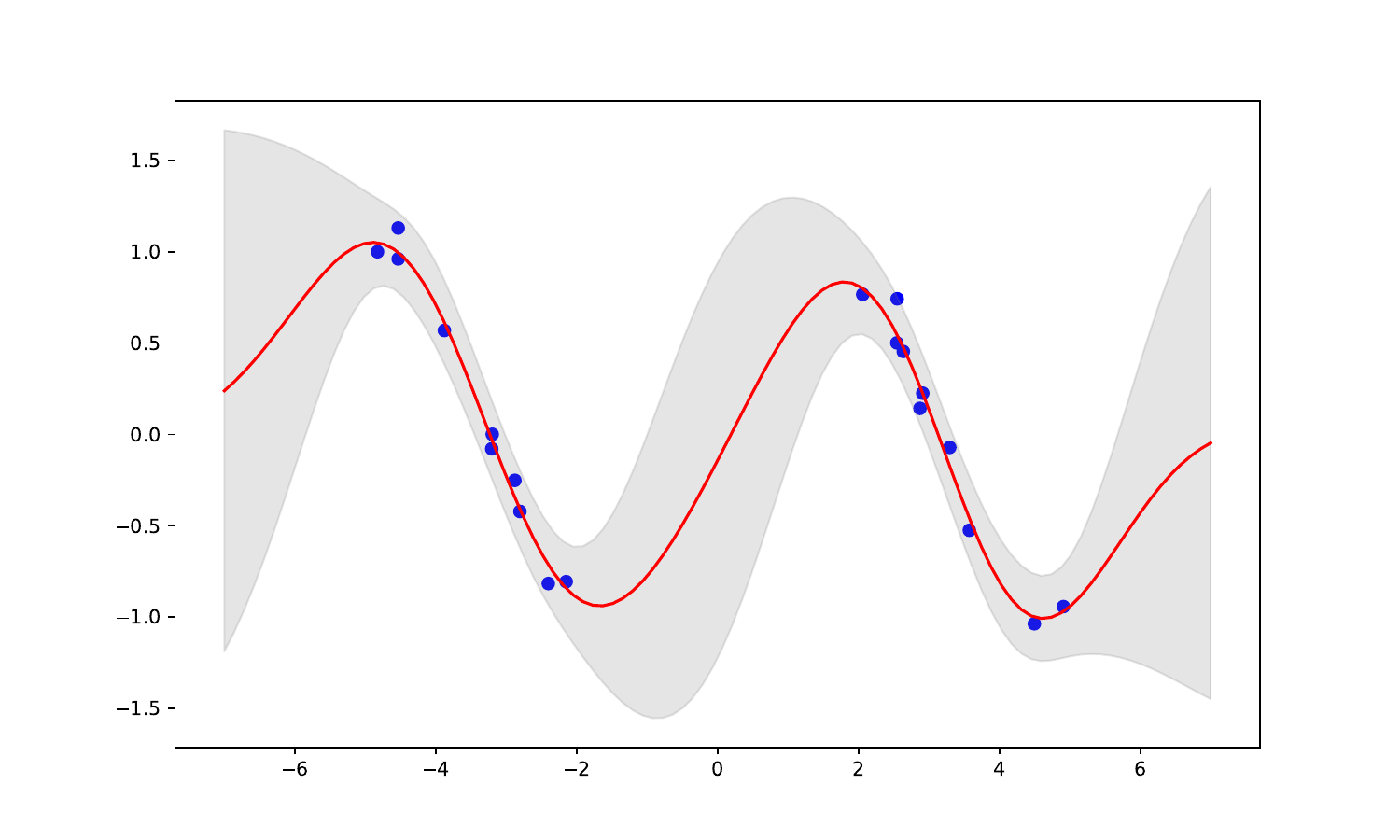}
    \caption{Simple example of Gaussian Process fit one-dimensional input space. The red line denotes the mean fit for the $f(x)$ function, while the grey bars denotes the 95\% confidence interval around $f(x)$. Blue dots depict training data points. Notice how the confidence is very high in regions dense with data points, while is very low in regions with no data points, demonstrating the desirable uncertainty-quantification properties of GPs.}
    \label{fig:1}
\end{figure}

\subsection{Gaussian Process Classifiers}
% In essence, 

Gaussian Processes (GPs) are flexible non-linear regression tools where a prior is placed on the functional form of a learner $f(x) \sim \mathcal{GP} \big( m(\cdot), K (\cdot, \cdot) \big)$, where $m(\cdot)$ is a mean function and $K(\cdot, \cdot)$ is a covariance or kernel function \cite{williams2006gaussian}. As is often the case, the prior mean $m(\cdot)$ is set to a constant, so that the kernel function $K(x, x') = \text{Cov} \big( f(x), f(x') \big)$ entirely determines the functional form of the learner, and the uncertainty around it, fitted based on the training data.

The choice of the kernel function $K(\cdot, \cdot)$ is thus crucial as it encodes our prior assumptions about the function we wish to learn (e.g., degree of smoothness). Popular kernel functions are the Squared Exponential (or Radial Basis Function) kernel, the Matérn kernel, and the periodic kernel, each imparting different functional properties such as smoothness, periodicity, and flexibility. We specifically pick what is perhaps the most popular and most flexible kernel, i.e., the Squared Exponential (SE) one:
\begin{equation*}
    K(x, x') \equiv \sigma^2_f \exp \left( -\frac{1}{2l^2} \| x - x' \|^2 \right)
\end{equation*}
The SE kernel has two main hyper-parameters that need to be optimized: i) the signal variance $\sigma^2_f$, which denotes the variance around the function $f(\cdot)$; ii) the length-scale $l$ which governs the smoothness of the functional approximation (i.e., how ``quickly" the function changes from input to input). These hyper-parameters are optimized via maximization of the marginal log-likelihood $\log p(y | l, \sigma^2_f)$. 

% Variability in predictions falls out of this process as the foundational building blocks of GPs are Normal Distributions. More specifically, given \textit{n} data points \((x_1, y_1), \ldots, (x_n, y_n)\), where \( y \in \mathbb{R} \) and \( x \in \mathbb{R}^d \), a GP regression models a relationship  \( y(x) \colon x \to y \) in the following way:
% \newline
% A pairwise kernel function \( K(x, x') \) between any two points \( x \) and \( x' \) in \( \mathbb{R}^d \) is chosen for the data to encode our beliefs about the covariance of two arbitrary data points. The usual choice of kernel \( K(x, x') \) is the Radial Basis Functions (RBF) denoted:

% The RBF kernel can be understood simply as representing a strong positive correlation if \( x \) and \( x' \) are in proximity to each other.
% A Gaussian Process is defined to be the random function \( y(x) \sim \mathcal{GP}(0, k(x, x')) \) such that for any \( n \) data points, \( (y_1, \ldots, y_n) \) is Gaussian distributed with \( n \times n \) covariance matrix \( K_n \) where \( K_n(i, j)\) is our kernel  \(k(x_i, x_j)\) .

Once the GP has trained on a set of inputs $X$ and corresponding outputs $y$, we can make predictions at new test points $X^*$. The joint distribution of the observed outputs \( y \) and the predicted outputs \( f^* \) is given by:
\[
\begin{bmatrix}
y \\
f^*
\end{bmatrix}
\sim \mathcal{N} \left(
\begin{bmatrix}
\mu \\
\mu^*
\end{bmatrix},
\begin{bmatrix}
K_n + \sigma_n^2 I & K_{n,m} \\
K_{m,n} & K_{m,m}
\end{bmatrix}
\right),
\]
where \( K_n \) is our \(n \times n\) covariance kernel for the training points, \( K_{n,m} \) is the \(n \times m\) cross-covariance matrix between the training and test points, \( K_{m,n} \) is its transpose, and \( K_{m,m} \) is the \(m \times m\) covariance kernel of the test points. Here, \( \sigma_n^2 I \) represents exogenous, additive, mean-zero noise around the output $y$, such that: $y = f(x) + \varepsilon$ where $\varepsilon \sim \mathcal{N} (0, \sigma^2_n)$.
The posterior distribution of the function values at the test points \( f^* \) given the observed data \( y \) is then:
\[
f^* | X^*, X, y \sim \mathcal{N} (\bar{f}^*, \text{cov}(f^*)),
\]
where
\[
\bar{f}^* = K_{m,n} (K_n + \sigma_n^2 I)^{-1} y
\]
and
\[
\text{cov}(f^*) = K_{m,m} - K_{m,n} (K_n + \sigma_n^2 I)^{-1} K_{n,m}.
\]

\begin{figure}[t]
    \centering
    \includegraphics[width=0.48\textwidth]{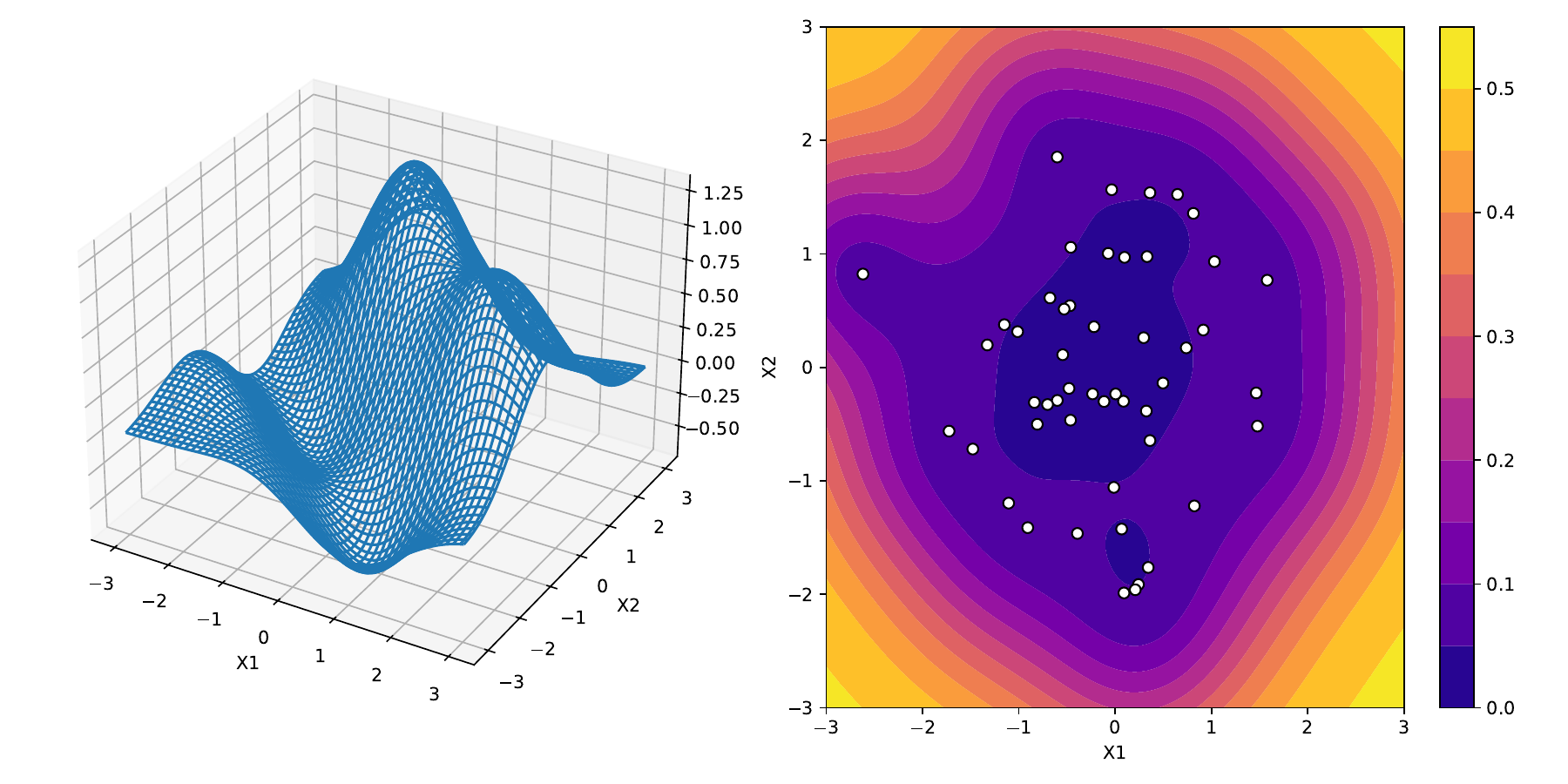} 
    \caption{Example of Gaussian Process fit on a two-dimensional inputs case. The 3D plot on the left depicts the mean fit for $f(x_1, x_2)$, while the contour plot on the right depicts the variance around $f(x_1, x_2)$. Variance is higher (brighter colour) in regions at the corners with fewer data points.}
    \label{fig:GP_2}
\end{figure}

Figures ~\ref{fig:1} and ~\ref{fig:GP_2} illustrate the use of GP regression models on simple one-dimensional and two-dimensional input examples, respectively. These toy examples are very useful to demonstrate the capabilities of GPs in terms of flexible function approximation and uncertainty quantification to efficiently detect out-of-sample regions. Similarly to the case of neural networks, it is then possible to easily extend the GP framework to adapt to classification problems with discrete outputs, such as the ones we will encounter in the experimental section. This is done by applying a sigmoid function to squash the `latent' output in the range $[0, 1]$, i.e.,
\begin{equation*}
    p(y = 1 | x) = \sigma(f(x)) = \frac{1}{1 + \exp(-f(x))}
\end{equation*}
in the case of two classes and binary output.

% We can easily convert a GP Regression model into a multi-class classifier by restricting its output to a number of levels equivalent to the number of classes. This approach gives similar results to a conventional multinomial logistic regression model, with the added benefit of knowing the variance in each prediction. For example, using a GP Regression model as a binary classifier based on a singular input may look like Figure ~\ref{fig:3}

% \begin{figure}[h!]
%     \centering
%     \includegraphics[width=0.5\textwidth]{figs/GP_Classifier.pdf} 
%     \caption{Binary GP Classifier}
%     \label{fig:3}
% \end{figure}

Gaussian Processes have demonstrated performance comparable to neural networks in many tasks, which is not surprising given that a GP can be viewed as a one-layer neural network with infinite width \cite{neal1996priors, lee2017deep}. However, GPs retain a significant advantage in their ability to provide robust uncertainty quantification. This makes them particularly valuable in applications where understanding the confidence of predictions is paramount, such as cyber-security ones.

Despite these strengths, the primary challenge with GPs lies in their scalability. The computational complexity of GPs amounts to $O(n^3p^3)$ with respect to the number of data points $n$ and input dimensions $p$. This cubic scaling can become prohibitive for large datasets or high-dimensional problems. To alleviate this issue in online classification settings, we will explore the use of a rolling buffer approach. The rolling buffer method allows for efficient updates and predictions while maintaining a manageable computational load. It's worth noting that scalability concerns are not unique to GPs; neural networks can also face challenges with very large online datasets, although their scaling properties are better.

Lastly, similar to neural networks, standard GPs are typically considered "black-box" models, lacking inherent interpretability. While they provide excellent predictions and uncertainty estimates, understanding the underlying decision-making process can be challenging. This limitation necessitates the use of post-hoc explanation techniques to gain insights into the model's behaviour and decision rationale.

% Of course, a similar classifier can be made by the conventional neural network approach and in fact, an equivalence between infinitely wide neural networks and Gaussian Processes has been shown by Lee et al.~\cite{lee2018deepneuralnetworksgaussian}. 

% Using GP models as predictive models such as classifiers has the advantage that the variance of a prediction is not approximated but falls out of our model. In other words, we always know precisely how confident our model is about a decision. 
% Despite these benefits, Gaussian Processes have one major issue, their scalability. Given the nature of performing the necessary posterior calculations in Gaussian Process regression, we must compute the inverse kernel matrix which gives such models a time complexity of \(O(n^3)\).

\section{Generalized Additive Models}

Generalized Additive Models (GAMs) are a flexible class of statistical models that extend generalized linear models by allowing for nonlinear relationships between predictors $x$ and the response variable $y$, while maintaining inherent interpretability thanks to their functional form assumptions \cite{hastie2017generalized}. A GAM assumes that the expected value of $y$, denoted as $\mathbb{E}[y]$, is related to the predictor variables through a link function $g(\cdot)$ and a sum of smooth functions as follows:
\begin{align} \label{eq:GAM}
g(\mathbb{E}[Y]) & = \beta_0 + \sum^p_{j=1} f_j (x_j) = \\ 
& = \beta_0 + f_1(x_1) + f_2(x_2) + ... + f_p(x_p) ~ , \nonumber
\end{align}
where $g(\cdot)$ is a specified link function (e.g., identity for normal distribution, logit for binary outcomes), $\beta_0$ is the intercept, and
$f_j(\cdot)$ are input-specific smooth functions. The smooth functions $f_j(\cdot)$ are typically represented using basis expansions:
\begin{equation*}
f_j(X_j) = \sum_{k=1}^{K_j} \beta_{jk} b_{jk}(X_j)
\end{equation*}
where $b_{jk}(\cdot)$ can be, e.g., cubic splines, thin plate regression splines, etc., and $\beta_{jk}$ are scalar coefficients. The model is estimated by maximizing a log-likelihood criterion.

GAMs represent a `silver bullet' within the range of statistical regression models, serving as a bridge between straightforward, easily interpretable linear models and more intricate, often less transparent machine learning algorithms. Their importance stems from a unique combination of inherent interpretability and predictive power, making them invaluable in many applied settings. The input-specific functions $f_j (\cdot)$ have an intuitive and straightforward interpretation as \emph{Shapley values} \cite{shapley:book1952}, and allow practitioners to visualize and understand the individual contribution of each variable to the response. Finally, GAMs do not generally compromise on predictive power significantly. In many real-world problems, they can achieve predictive accuracy comparable to more complex, black-box models such as random forests or neural networks \cite{rudin2019stop}.

Finally, it is worth reminding that, in case the expressivity of the GAMs specification (\ref{eq:GAM}), in terms of the complexity of the functions they can approximate, is believed not to be enough for the problem at hand, one can resort to the GA$^2$M extension, that reads:
\begin{equation*}
    g(\mathbb{E}[Y]) = \beta_0 + \sum^p_{j=1} f_j (x_j) + \sum^p_{j=1} \sum^p_{j'>j} f_{j j'} (x_j, x_{j'}) ~ .
\end{equation*}
This specification contains additional basis expansion components capturing pairwise feature interactions $f_{j j'} (\cdot, \cdot)$.

\begin{figure}[t]
    \centering
    \includegraphics[width=0.44\textwidth]{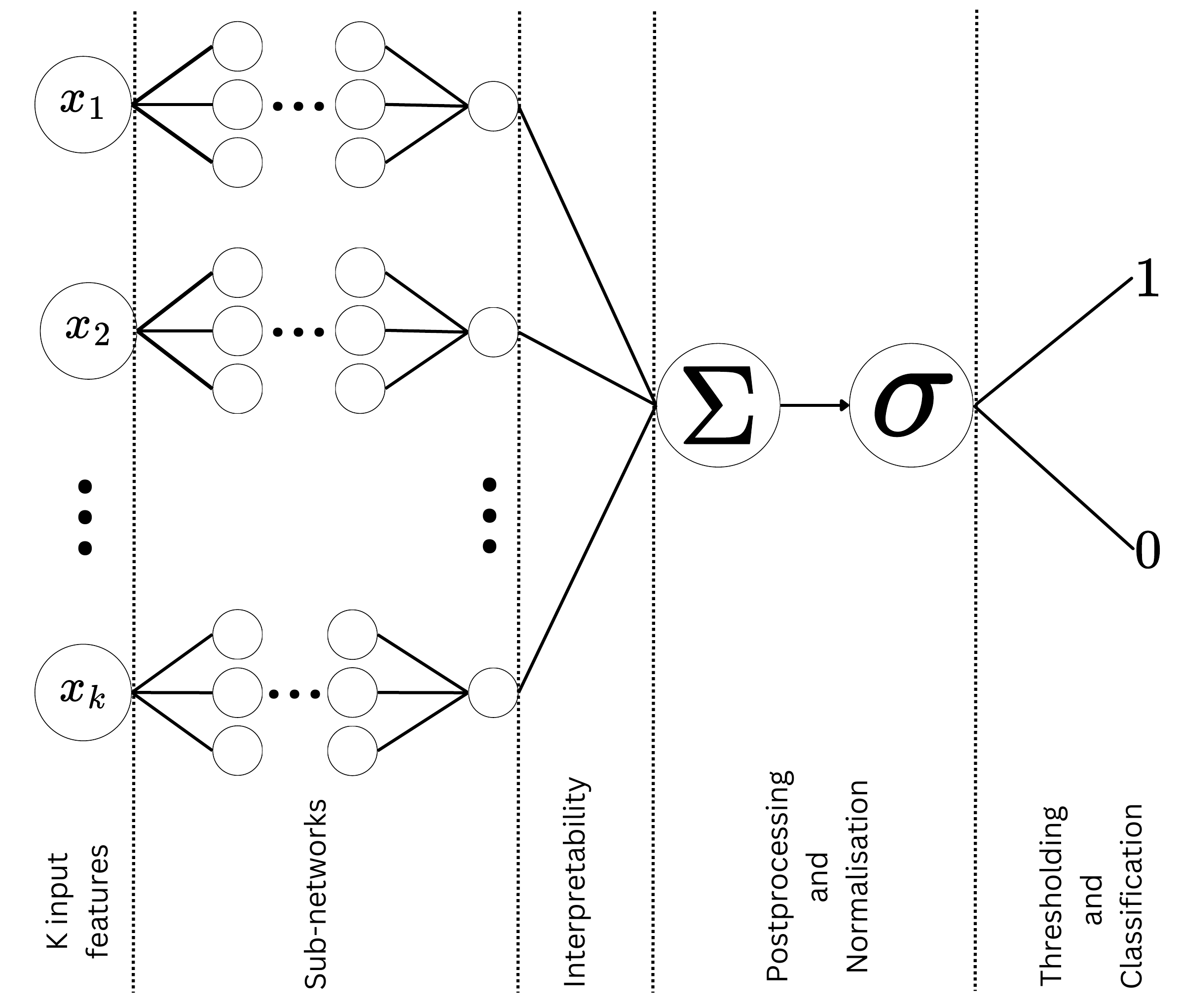} 
    \caption{Architecture of a NAM: each input is modelled via an input-specific fully-connected MLP (sub-networks) that guarantee interpretability of their contribution $f_j(\cdot)$ (Shapley value \cite{shapley:book1952}) to the final output $y$. Their functions are then summed up and normalized to construct the final predictor.}
    \label{fig:4}
\end{figure}

\subsection{Neural Additive Models}

Neural Additive Models (NAMs) were first proposed in \cite{agarwal2021neural} and fall within the GAMs class. They introduce restrictions in the conventional neural network architecture and model each input-specific function $f_j (\cdot)$ via a separate neural net (sub-network) which is then summed up as described in Eq.~\ref{eq:GAM}, instead of modelling the output $y$ as a fully-connected Multi-Layer Perceptron (MLP) $f(\cdot)$. This allows one to harness the inherent interpretability of GAMs while retaining all the nice computational properties of MLP training, i.e., back-propagation, mini-batching, parallelization, etc. Figure \ref{fig:4} reports a visual representation of a NAM architecture for classification tasks, together with a description of each component in the structure.

Despite their advantages, NAMs lack built-in uncertainty quantification, a key feature that is inherent to Gaussian Processes (GPs) instead, as they carry the same over-confidence issues typical of deep learning architectures \cite{guo2017calibration, lakshminarayanan2017simple, rahaman2021uncertainty}. This can be a significant limitation in high-stakes applications. We will discuss in the next sections how both advantages brought about by GPs and NAMs (or GAMs more generally) can be harnessed together, by imposing an additive kernel GP structure that can be proved to yield the GAMs form described in Eq.~\ref{eq:GAM}.

For comparison with the other uncertainty-aware GP methods in the online classification tasks featured in the experimental section, we will construct an estimator for the variance of the classification output $p_i$ of the neural nets methods (standard NN and NAM) as $p_i (1 - p_i)$, where $p_i = g \big( f (x_i) \big)$ and $g(\cdot)$ is the logit function.

\subsection{Additive Gaussian Processes}

Additive Gaussian Processes (AGPs) \cite{duvenaud2011additive} represent another approach within the GAMs framework, combining the flexibility and uncertainty-awareness of Gaussian Processes with the interpretability of additive models. Unlike NAMs which use neural networks for each $f_j (\cdot)$, AGPs model each component function as an independent Gaussian Process $f_j \sim \mathcal{GP} \big( 0, k_j(x_j, x'_j) \big)$, where $k_j$ is a kernel function specific to the $j$-th dimension, via an aggregate kernel function of the form:
\begin{equation*}
    k(x, x') = k_0 + \sum_{j=0} k_j(x_j, x'_j) ~ .
\end{equation*}
Initialising the kernel in this manner enables us to zero out all the dimensions other than \(x_j\) in order to find the individual, input-specific, 1-D GP Classifier relating to \(x_j\), for all \(j \in [1, p] \subset \mathbb{N}\). Since the individual kernels \(k_j\) are pairwise independent this pipeline is equivalent to learning \(k\) 1-D GP Classifiers for each feature of the data, however, it is generally computationally faster for the model to learn the first.

\begin{figure}[t]
    \centering
    \includegraphics[width=0.45\textwidth]{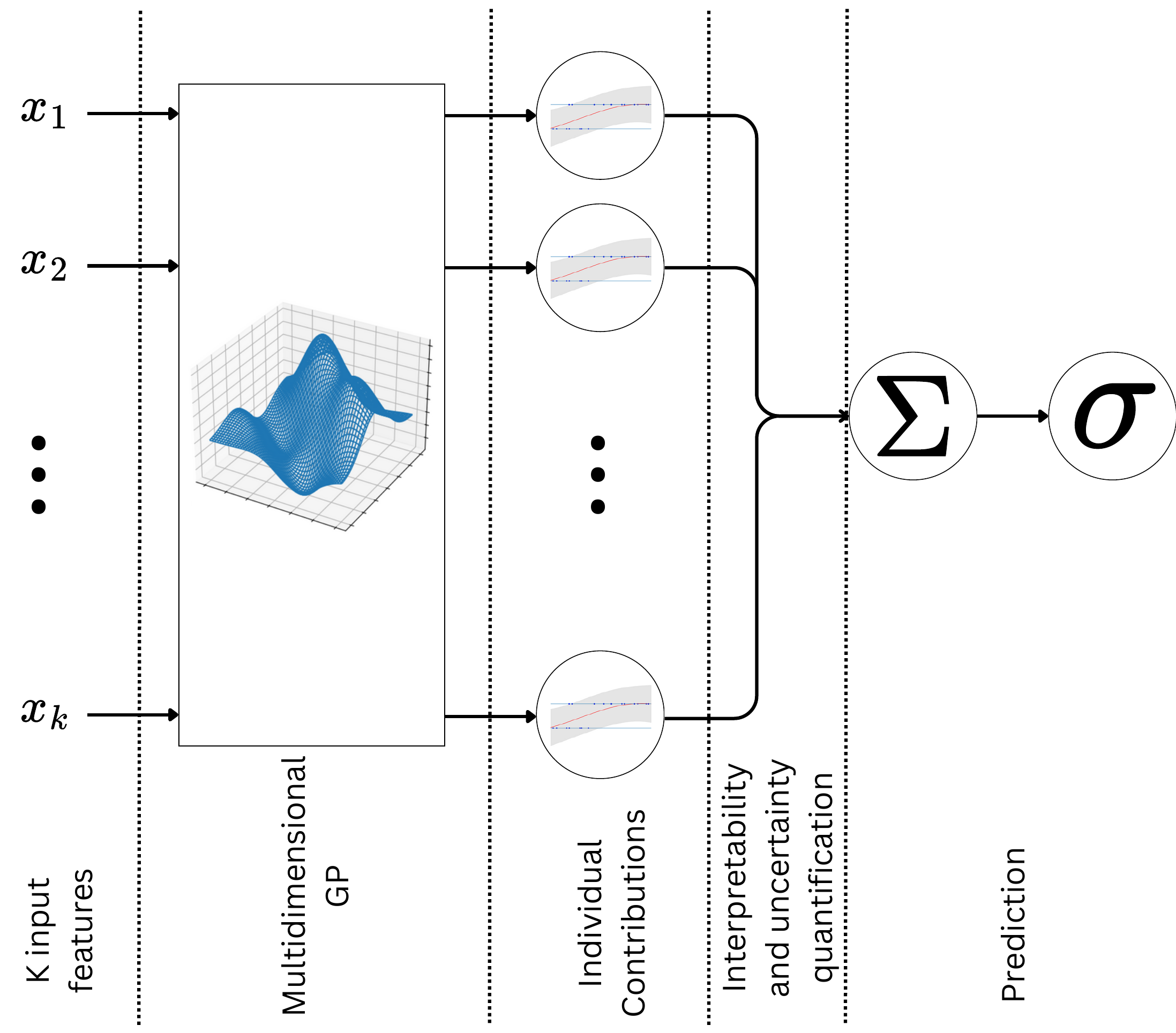} 
    \caption{Architecture of an Additive GP: each data feature is fed into a Multidimensional GP, then individual contributions are extracted, summed and normalised.}
    \label{fig:5}
\end{figure}

AGPs maintain the interpretability advantages of GAMs and NAMs while introducing built-in uncertainty quantification, addressing a key limitation of NAMs mentioned earlier. Like the input-specific functions in GAMs and NAMs, each $f_j(x_j)$ in AGPs can be interpreted as a Shapley value, representing the individual contribution of each variable to the response. This allows for clear visualization and understanding of feature-specific effects, similar to the benefits offered by NAMs. Furthermore, in the case of Additive GPs it is easier to specify GA$^2$M additional feature interactions compared to NAMs, as this is achieved by simply modifying the kernel's additive structure (see \cite{duvenaud2011additive}). In contrast, feature interactions in NAMs must be manually created and fed into the architecture as additional features. Figure \ref{fig:5} above contains a visual depiction of an Additive GP structure.

Despite their advantages, AGPs still face scalability challenges, struggling with large sample sizes due to their $\mathcal{O}(n^3)$ computational complexity (per input, i.e., $p=1$). While GP-NAM \cite{zhang2024gaussianprocessneuraladditive} could potentially alleviate this computational burden through its random Fourier feature approximation, it would sacrifice the full Gaussian Process formulation and, consequently, some of the uncertainty quantification benefits. To address this scalability issue in online learning problems, while maintaining the benefits of full GPs, we propose using a rolling buffer that cleverly samples from past collected experience, allowing for efficient computation without compromising on the model's uncertainty awareness.

\subsection{Scalability Issues}

The foundational cause of the computational inefficiency of both GPs and Additive GPs is their requirement to compute the inverse of the covariance matrix kernel which is then used for posterior calculations. Inefficient and long-runtime models are unlikely to be adopted in real-life applications, especially in high-stakes environments such as cyber-security. In these settings, large, feature-heavy datasets are typically used to provide models with as much context as possible to minimize risk.

Scalability issues can be further exacerbated in online learning settings since as the model encounters new batches of data it necessitates frequent re-training to take into account local and global distributional shifts in the data generating process of $p(y,x) = p(y|x) p(x) = p(x|y) p(y)$. These can generally occur in either the marginal input distribution $p(x)$ (also known as `covariate shifts'), or in the marginal output distribution $p(y)$ (known as `semantic shifts') \cite{yang2024generalized}. Routine updates and re-training of the models are then necessary to adjust predictions to these shifts happening over time and avoid biases.

\vspace{0.2cm}

\textbf{Sample Window.} In some online settings re-training each time on the whole dataset is simply prohibitive, for any model. As such, we propose using a rolling buffer that considers re-training on a sub-sample of the whole dataset made of the most recent batches of new data, plus randomly drawn batches of historical data. We demonstrate how this strikes a good balance between: i) computational feasibility; ii) the need to adjust to local drifts/shifts occurring in the data generating process; iii) and the need to learn global patterns as well. This enables us to employ GPs and Additive GPs more comfortably without running into major scalability issues, as well as other models such as NNs or NAMs, although they do scale better as discussed earlier. An alternative method to improve GPs scalability is represented by Sparse Input Gaussian Processes \cite{quinonero2005unifying} which learns an optimal sub-sample of `inducing' points. However, learning the set of inducing points at every re-training step introduces some computational burden. In addition, this would not guarantee that the model focuses on the latest trends in the new batches of data. Obviously, the ratio of recent to random historical data in the proposed buffer is a problem-specific hyper-parameter that can be tuned considering the computational complexity and accuracy trade-off. Online learning problems where distributional drifts over time are believed to be smooth do not perhaps need as many recent samples (and perhaps as many re-training steps) as historical ones to achieve the same level of accuracy, and vice versa.

\vspace{0.2cm}

\textbf{Active Learning.} In online learning cyber-security problems, it is often necessary to manually label the new inputs before being able to incorporate them into the model and re-train (e.g., malware versus goodware). However, this labelling process can be time-consuming and expensive, especially given the large volumes of data typically involved in cyber security applications \cite{sommer2010outside}. In order to tackle this challenge, practitioners typically resort to active learning techniques. Active learning allows for selective labelling of the most informative samples from the new batch, thereby reducing the overall labelling cost while maintaining or even improving model performance \cite{settles2009active, ren2021survey}. The selection of these samples is typically guided by ``acquisition functions" that aim to identify the most valuable instances for labelling. Some examples of popular active learning acquisition strategies are based on uncertainty sampling \cite{hanneke2014theory}, expected error reduction \cite{roy2001toward} and query by committee \cite{seung1992query}. The first category, `uncertainty sampling' is arguably the most popular one, as it is as effective as intuitive to implement and deploy. However, for this technique to work best, we need to ensure that the model is capable of accurately quantifying uncertainty around the new samples. When using models with strong uncertainty quantification properties, such as Gaussian Processes (GPs) or Additive GPs, we can leverage more sophisticated acquisition functions based on Information-Theoretic criteria. These functions are particularly effective because they can target samples for which the model's epistemic uncertainty is high. Epistemic uncertainty refers to the model's uncertainty due to lack of knowledge or information about certain scenarios, as opposed to aleatoric uncertainty, which represents inherent randomness in the data that cannot be reduced by collecting more information \cite{der2009aleatory}. This distinction is crucial, as epistemic uncertainty can be reduced by gathering more data or improving the model, making it particularly relevant for active learning, while aleatoric uncertainty represents irreducible noise in the data. For instance, in malware detection, epistemic uncertainty might arise from the model's lack of exposure to a new type of malware, while aleatoric uncertainty could stem from inherent randomness in certain features of the software \cite{kendall2017uncertainties}. While NNs and NAMs can be powerful predictive tools, they typically lack built-in mechanisms for quantifying uncertainty, especially for distinguishing between epistemic and aleatoric uncertainty. On the other hand, GPs (and Additive GPs) are inherently well-suited to tell the two types of uncertainties apart.

% In some sequential decision-making problems, a verification step can occur after the model makes a decision on unseen data, allowing for corrections, typically by a human operator, if the model is wrong. However, this correction process often incurs intrinsic costs, such as human labour, and can become especially burdensome when the model must make numerous rapid decisions in an online setting where the volume and speed of predictions exceed the capacity for complete verification. To address this, we can utilize \textit{active learning}, where the verification step prioritizes correcting decisions with the highest uncertainty within the constraints of available resources. This approach mitigates the performance loss associated with using the window by focusing corrections on the decisions where the model's confidence is lowest.

\begin{figure}[t]
    \centering
    \includegraphics[width=0.45\textwidth]{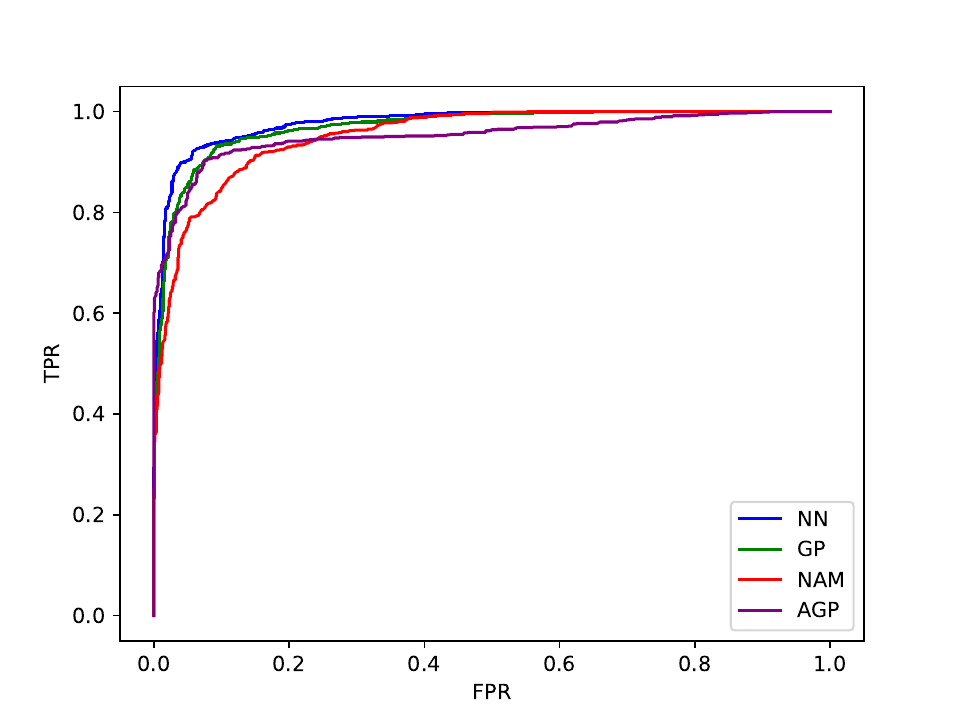} % Replace 'example-image' with the path to your image file
    \caption{ROC curves of the four models considered: Neural Nets (NN), Gaussian Processes (GP), Neural Additive Models (NAM) and Additive GP (AGP). We used a window of 20\% (of the data) for routinely re-training the models.}
    \label{fig:6}
\end{figure}

\section{Experiments}

In this section, we present a comprehensive empirical evaluation of Neural Nets and GPs, and their interpretable counterparts, NAMs and Additive GPs, on the online learning problem of URL phishing classification. We showcase the desirable properties of NAM and Additive GPs in terms of the inherent interpretability of the model's decision-making process, and the uncertainty quantification capabilities of GPs.

\subsection{Dataset and Problem Setup}

To demonstrate the applicability of the proposed models in online learning settings, we utilize a tabular dataset of URLs\footnote{\url{https://www.kaggle.com/datasets/eswarchandt/phishing-website-detector}}, containing over 11,000 URLs scraped from the Internet. Each sample in this dataset is characterized by 30 features $x \in \mathcal{X}$, along with a binary classification label $y \in \{0,1\}$ indicating whether the URL is malicious (phishing) or genuine.

To closely simulate an online sequential decision-making environment, which is typical in real-world cyber-security applications, we configure the following experimental setup:

\begin{enumerate}[topsep=0pt,itemsep=0ex,partopsep=0ex,parsep=0ex, leftmargin=4ex]
\item Initial Training: the model is trained on an initial window of the URL training data of predetermined size, which represents a pre-training offline dataset to warm up the model.
\item Sequential Updates: a random batch of new URLs, containing between 50 to 100 samples, is used to update the model. 
\item Active Learning (Optional): If active learning is employed, the predicted classes are ranked by the model's predictions' variance, and a pre-specified number of new inputs are selected to reveal the corresponding labels to update the model.
\end{enumerate}

\subsection{Baseline and Comparisons}
Throughout the experiments, we evaluate four different models to provide a comprehensive comparison:

\begin{figure}[t]
    \centering
    \includegraphics[width=0.46\textwidth]{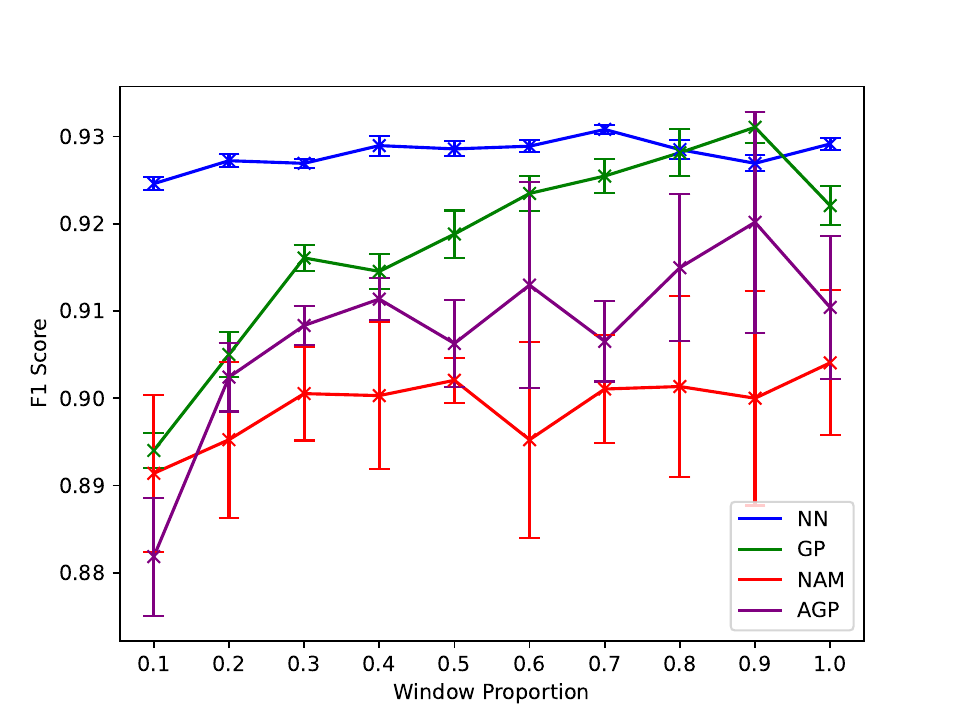} % Replace 'example-image' with the path to your image file
    \caption{F1 classification scores of the four models (NN, GP, NAM and AGP), for increasing window proportion values.}
    \label{fig:7}
\end{figure}

\begin{enumerate}[topsep=0pt,itemsep=0ex,partopsep=0ex,parsep=0ex, leftmargin=4ex]
\item \textbf{Neural Network (NN)}: Serves as the baseline for model comparisons. It represents a conventional choice for classification tasks and is equipped with estimated decision variance for uncertainty awareness, estimated by $p_i (1 - p_i)$, where $p_i$ is the conditional probability $p_i = p(y=1 |x)$.
\item \textbf{Gaussian Process (GP)}: A multidimensional full GP classifier that provides true decision variance, offering robust uncertainty estimation.
\item \textbf{Neural Additive Model (NAM)}: A NAM classifier that outputs estimated input $j$ specific contributions to the final classification $p_j$, that can be interpreted as Shapley values, thus offering inherent interpretability. However, similarly to NNs, they do not provide a measure of epistemic uncertainty.
\item \textbf{Additive GP (AGP)}: Our proposed Additive Gaussian Process classifier, which integrates the strengths of both GPs and NAMs in terms of uncertainty quantification and inherent interpretability, respectively. It provides input-specific contributions to the final classification and the true posterior variance around these contributions, which are useful for uncertainty-based sampling in active learning.\\
\end{enumerate}

Figure~\ref{fig:6} shows the Receiver Operating Characteristic (ROC) curves for each model, showing that they all perform suitably well, despite having a 0.2 'Window Proportion`:
\[
\text{Window Proportion} = \frac{\text{Window Size}}{\text{Training Data Size}} 
\]

\noindent The associated Area Under Curve (AUC) scores of the ROC plot are as follows: Neural Network (NN) achieved 0.978, Gaussian Process (GP) 0.955, Neural Additive Model (NAM) 0.969, and Additive GP (AGP) 0.952. These results indicate that all models perform well, with no largely statistically significant difference. The Neural Net, perhaps unsurprisingly, turns out to be the most accurate model. However, it has the disadvantage of being entirely black-box. While NAM and AGP fall a bit behind, they sacrifice not too much accuracy to guarantee interpretable outputs, and in the case of AGPs, also good uncertainty quantification around the interpretable outputs.

% These metrics are supported by the associated ROC Area Under Curve scores:
% \newline
% \begin{center}
% \begin{tabular}{||c c c c||} 
%  \hline
% NN & GP & NAM & GP-NAM \\ [0.5ex] 
%  \hline§
%  0.978 & 0.955 & 0.969 & 0.952 \\ 
%  \hline
% \end{tabular}
% \end{center}

\subsection{Sample Window and Active Learning}
We further explore the impact of our proposed modifications to the online learning pipeline on each model's performance when we consider a sub-sample window of the entire dataset to guarantee better scalability when re-training the models, and when we employ clever active learning sampling strategies.

Figure~\ref{fig:7} above depicts the change in F1 classification scores for increasing window proportion values, for each of the four models considered. Unsurprisingly, all models perform significantly better when a larger window is used, as this allows the model to be trained on most of the data. However, this comes at a significantly higher computational cost of re-training for all four models, especially for the two GPs. Overall, we can observe though that for smaller proportions of data (e.g., around 20\%-30\%) being used for re-training, the models do not incur significant losses in performance, and still achieve good overall accuracy.

\begin{figure}[t]
    \centering
    \includegraphics[width=0.47\textwidth]{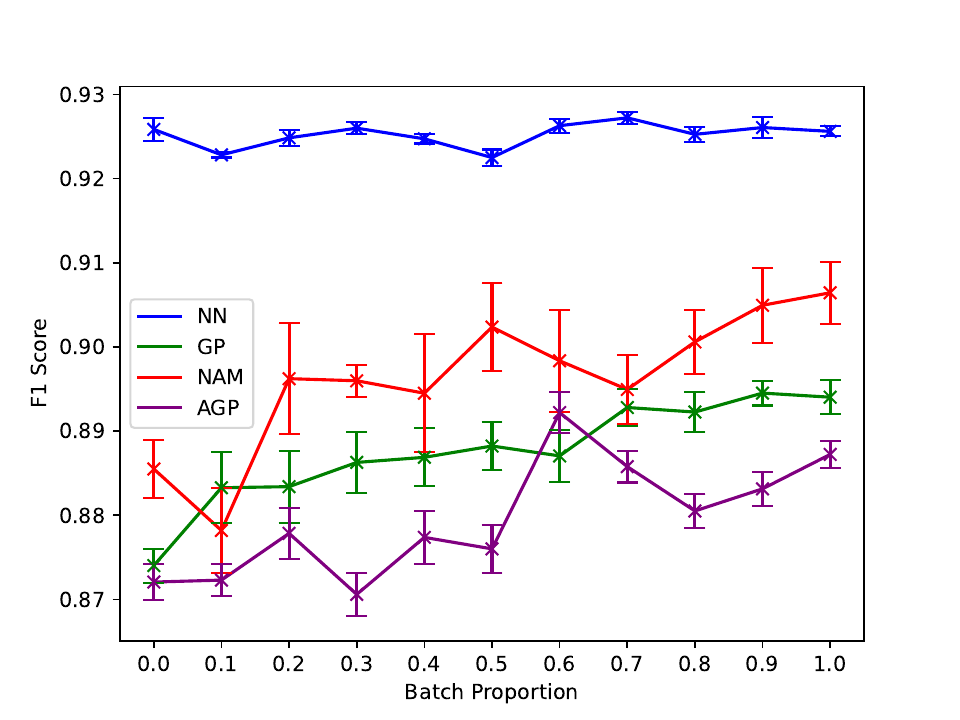} % Replace 'example-image' with the path to your image file
    \caption{F1 classification scores of the four models (NN, GP, NAM and AGP), for an increasing proportion of labelled data in the new batches.}
    \label{fig:8}
\end{figure}

Figure~\ref{fig:8} instead demonstrates what happens if we increase the proportion of the data points in the new batches of inputs that we label via active learning acquisition strategy. The acquisition strategy we use to select the data points to be labelled and used to re-training in the new batch is based on uncertainty sampling as discussed earlier, i.e., we select the points for which the model's prediction is most uncertain. In general, we observe a slight upward trend in the models' F1 scores if we increase the proportion of newly labelled samples to 100\%. However, this does not bring about massive improvements in performance for any of the compared models. This hints at the fact that the uncertainty-based active sampling strategy is able to select highly representative samples to be labelled, even in very small proportion values (i.e., 10\%-20\%). Overall this implies that, considering that labelling new data is often a costly process, using only a small proportion of newly labelled samples selected via uncertainty-based active learning is sufficient to achieve a desirable performance level during re-training.

\subsection{Interpretability and Uncertainty-Awareness}

Having established the performance of our models in the URL phishing classification problem, we now explore the power of interpretability and uncertainty-awareness in this context. Our focus will be on the two interpretable models: Neural Additive Model (NAM) and Additive Gaussian Process (AGP).

\begin{figure}[t]
    \centering
    \includegraphics[width=0.47\textwidth]{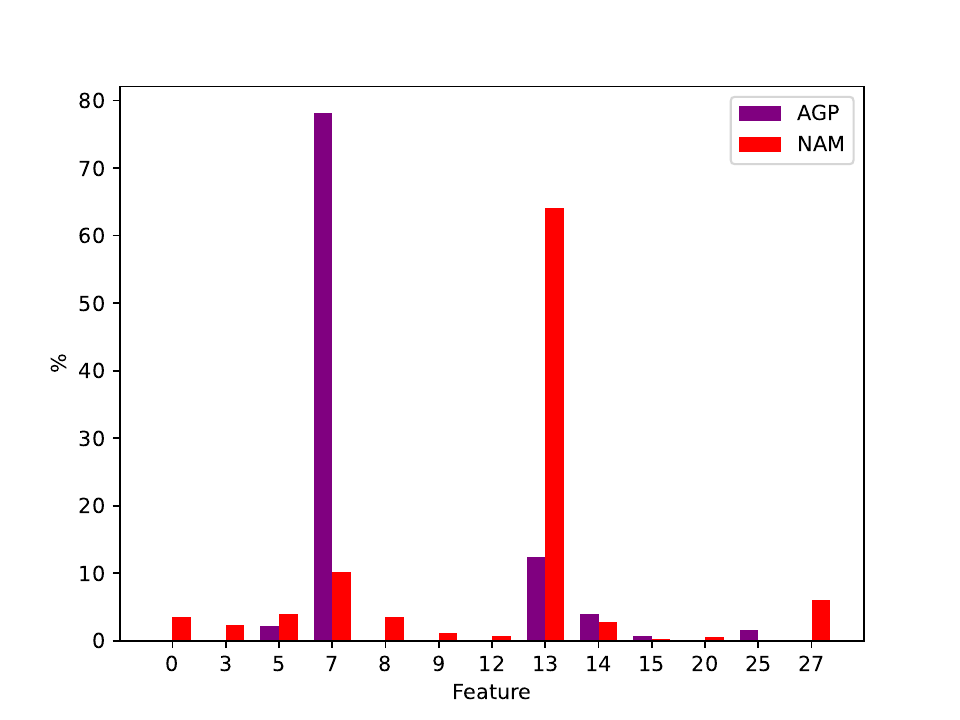} % Replace 'example-image' with the path to your image file
    \caption{Percentage of times a certain feature $x\in\mathcal{X}$ from the dataset gets picked by the models as the main contributor to the final classification. Feature 7 and 13 are the most important ones picked by both models and correspond to ``URL not being HTTPS" and ``URL not being an Anchor URL", respectively.}
    \label{fig:9}
\end{figure}

In Figure \ref{fig:9}, we report the proportion of times a certain feature $x_j \in \mathcal{X}$ from the dataset is picked by the model as the main contributor to the final classification of phishing. For clarity, we do not print out the whole 30 features in the dataset but report only the ones with higher contribution percentages. Note that most features out of the 30 are almost redundant, as they get picked very few times by the model (approximately $< 1\%$ of the times). Instead, we can observe that both NAM and AGP seem to picking up the same two features, number 7 and 13, as the most important ones for the final classification. These features refer to ``URL not being HTTPS" and ``URL not being an Anchor URL", respectively, which are sensible things to look at when trying to distinguish phishing emails.

% We observe that the GP-NAM considers feature 7 to be the largest contributor to a classification of phishing the vast majority of the time, feature 13 the next most frequent, while for the NAM it's the other way around. Either way, both interpretable models consider features 7 and 13 of the URL phishing dataset to contribute most to the classification of phishing in the vast majority of cases. Looking at the dataset we see that feature 7 refers to the URL not being HTTPS and feature 13 refers to the URL not being an Anchor URL\footnote{In initialising the models a classification of phishing was associated with a model output of '0', while the database describes a HTTPS Anchor URL through a values of '1` the respective feature columns, so it follows that if feature 7, for example, was the largest contributor to a classification of phishing it was also the smallest feature specific output in the model, referring to a small value in the feature 7 column of the dataset, which is associated with a URL not being HTTPS.}, which are exactly the key indicators that we would look for in identifying whether a URL is phishing or genuine, a great sanity check.

\begin{figure}[t]
    \centering
    \includegraphics[width=0.47\textwidth]{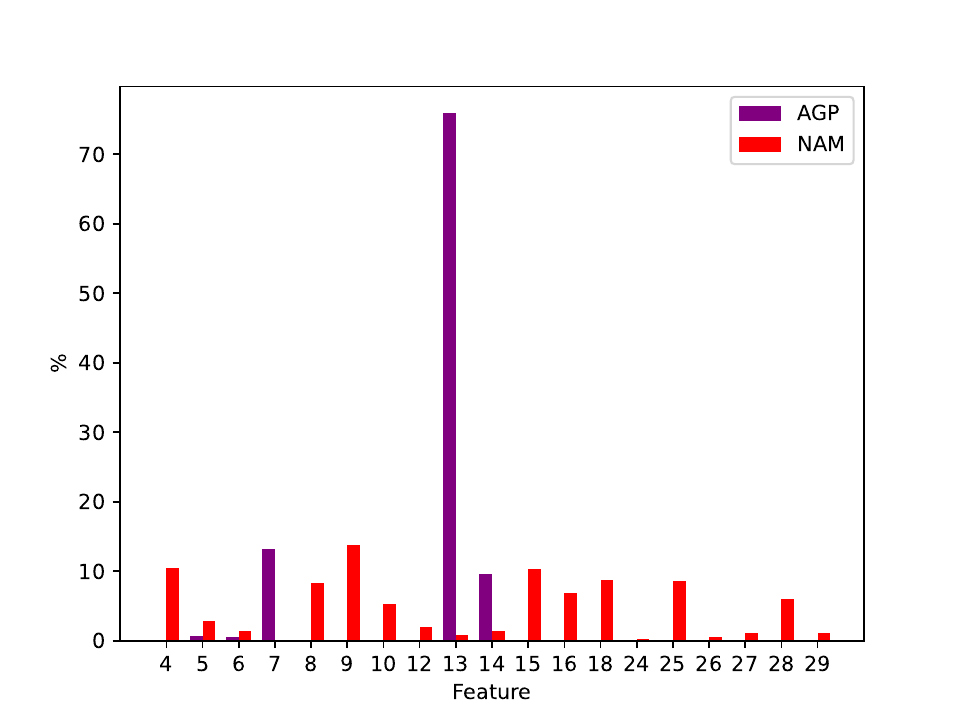} % Replace 'example-image' with the path to your image file
    \caption{Percentage of times a certain feature $x \in \mathcal{X}$ from the dataset gets picked by the models as the feature with the highest estimated variance of classification contribution. AGP selects feature 13, ``URL not being an Anchor URL", as the most variable one most of the time, while NAM seems not to be able to distinguish whether any of the features stand out in terms of contribution variance.}
    \label{fig:10}
\end{figure}

We then investigate the output of each of the two interpretable models (NAM and AGP) in terms of feature-specific uncertainty quantification. Figure \ref{fig:10} reports the percentage of times a specific input feature $x_j$ from the dataset emerges as the one with the highest variance around its predictive contribution $f(x_j)$ to the final classification. As in the case of the earlier plot, we report percentages for only a subset of the 30 features for clarity, excluding the ones with very low percentages. We can observe that AGP seems to be placing most of the model uncertainty around the classification contributions deriving from feature 7, ``URL not being HTTPS", and particularly feature 13, ``URL not being an Anchor URL". It is reassuring to know that the AGP model does not utilize feature 13 for classification decisions as much, as depicted in Figure \ref{fig:9}, due to its high variance. On the contrary, the NAM model seems not to be able to pinpoint exactly where most of its uncertainty lies, in terms of feature-specific variance, as it picks very different features each time as the most variable ones. This can be related to the phenomenon of over-confidence \cite{guo2017calibration, lakshminarayanan2017simple, rahaman2021uncertainty} typical of deep learning models, where the neural nets essentially ``does not recognize what it does not know". This implies that NN models end up struggling to detect out-of-sample data points, i.e., data points characterized by input values $x \in \mathcal{X}$ that the model has not seen during training time. This occurs generally for deep learning models, and thus for Neural Additive versions of them as well.

% \newline
% \newline
% It is interesting that both features 7 and 13, relating to a HTTPS Anchor URL appear most frequently in our interpretability and feature uncertainty plots for the GP-NAM, we believe that this behaviour sheds light on a better understanding of our model outputs and shows the power of such interpretable, uncertainty-aware models in more general high-stakes scenarios. This is demonstrated through the following example:
% \newline
% \newline
% \textit{The GP-NAM classifies a URL as phishing and considers feature 13 the biggest contributor to this decision. However, feature 13 is also the most variable a vast majority of the time, leading to the conclusion that we may be wary of such a classification given the cost of being incorrect set out in the high-risk problem.}
% \newline
% \newline
% We are attaining more information about our model's decisions than a traditional classification model with just its interpretability aspect, then only by incorporating uncertainty awareness and using both additional information sources in tandem do we ascertain a new level of confidence (or distrust) in our model's outputs.

\section{Related Work} 

In this section, we briefly discuss some related works on explainability/interpretability and uncertainty-quantification in cyber-security problems. Several contributions have previously addressed the challenge of explainable/interpretable models for cyber-security problems~\cite{charmet2022explainable, emerson2024cyborg++}, such as Intrusion Detection Systems (IDS). Among the most influential works,~\cite{abou2022should} provides an XAI-based framework for the ex-post interpretability of deep learning models in IoT intrusion detection.~\cite{wang2020explainable} specifically employs SHAP for local and global explanations for IDS decisions, while~\cite{mane2021explaining} showcases the use of SHAP, LIME and other methods.~\cite{marino2018adversarial} instead proposes an adversarial approach that uses feature modifications to explain incorrect classifications in IDS. As for the topic of uncertainty-quantification, early contributions such as~\cite{frigault2008measuring},~\cite{xie2010using} and~\cite{poolsappasit2011dynamic} demonstrate the need for uncertainty-aware probabilistic models such as Bayesian Network for cyber-security domains. Finally, there are very few recent contributions, such as~\cite{yang2024interpretable} and~\cite{yang2024towards}, at the intersection of interpretability and uncertainty-awareness in cyber-security ML models, where our work ultimately positions itself. Compared to these lines of work, our paper is the first that proposes a class of inherently interpretable, thus yielding faithful explanations by construction, and at the same time uncertainty-aware models, that are capable of quantifying epistemic uncertainty around new test data points and practically detect out-of-sample instances. We have demonstrated these capabilities specifically in online learning settings, which are abundant in the cyber-security domain~\cite{foley2022autonomous,foley2023inroads,caron2024view}, but these are extendable also to offline settings and sequential decision-making ones such as Reinforcement Learning.

\section{Conclusion}
In this paper, we addressed the need for interpretable and uncertainty-aware ML models in high-stakes cyber-security applications, particularly in online learning settings. We demonstrate that Additive Gaussian Processes (AGPs) offer a balanced solution, combining predictive performance, inherent interpretability, and robust uncertainty quantification. Our research tackles the scalability challenges of Gaussian Process-based models through a rolling buffer approach, significantly improving computational efficiency without compromising performance. Empirical evaluation on a URL phishing classification task showcases AGPs' practical applicability, providing actionable insights through feature-specific contributions and uncertainty estimates. The synergy between AGPs and active learning techniques optimizes data labelling processes, crucial in cyber-security applications where labelling is often costly. These findings have broad implications for machine learning in high-stakes environments, offering a tool that enhances AI trustworthiness and accountability while meeting regulatory demands for explainable AI.

While AGPs demonstrate compelling advantages in terms of uncertainty quantification and interpretability, it is still worth reminding that their inherent computational complexity remains a significant challenge and a trade-off, particularly in online learning settings where frequent model updates are necessary. Neural Additive Models (NAMs) were specifically included in the analysis to address these scalability concerns, leveraging the efficient training mechanisms of neural networks while maintaining the interpretability benefits of additive models. However, while NAMs offer improved computational efficiency and good predictive performance, they sacrifice some of the robust uncertainty quantification capabilities that make AGPs particularly valuable in high-stakes applications.

Future research directions include extending this framework to other domains other than the online learning one, such as the Reinforcement Learning one, where the need for interpretable autonomous cyber-defence agents is critical. By demonstrating the effectiveness of AGPs in online learning for cyber-security problems and addressing scalability challenges, we aim to encourage the adoption of transparent and reliable models in critical decision-making scenarios.

% Interpretable and uncertainty-aware models are crucial when addressing high-stakes problems such as finance, healthcare, and cyber-security. While GP-NAMs are among the leading models with these desirable properties, their standard versions face significant scalability challenges, limiting their adoption in these fields. By implementing a series of modifications to the training pipeline, we preserve the interpretability and feature uncertainty outputs of GP-NAMs. This achieved substantially improved scalability without a significant compromise in model performance as well as demonstrating the power of GP-NAMs over other interpretable models such as NAMs in online cyber-security problems, a venture GP-NAMs is novel too. 

% Interpretable, uncertainty-aware models are almost always required when working with any high-stakes problems: finance, healthcare or cyber-security. GP-NAMs are a top contender for models with these properties but their XXX have serious scalability issues which restrict their uptake in such scenarios. Through a series of modifications to their training pipeline, we maintain cohesive and useful interpretable and feature uncertainty outputs with notably better scaling without a significant loss in model performance. 

\section*{Acknowledgements}

Benjamin Kolicic was funded by the Alan Turing Institute’s Defence and Security partners. Alberto Caron, Vasilios Mavroudis, and Chris Hicks are funded by the Defence Science and Technology Laboratory (Dstl). The research supports the Autonomous Resilient Cyber Defence (ARCD) project within the Dstl Cyber Defence Enhancement programme.

\bibliographystyle{plain}
\bibliography{mybib}

\end{document}